# Texture-Based Input Feature Selection for Action Recognition

Yalong Jiang

*Abstract*—The performance of video action recognition has been significantly boosted by using motion representations within a two-stream Convolutional Neural Network (CNN) architecture. However, there are a few challenging problems in action recognition in real scenarios, e.g., the variations in viewpoints and poses, and the changes in backgrounds. The domain discrepancy between the training data and the test data causes the performance drop. To improve the model robustness, we propose a novel method to determine the task-irrelevant content in inputs which increases the domain discrepancy. The method is based on a human parsing model (HP model) which jointly conducts dense correspondence labelling and semantic part segmentation. The predictions from the HP model also function as re-rendering the human regions in each video using the same set of textures to make humans appearances in all classes be the same. A revised dataset is generated for training and testing and makes the action recognition model exhibit invariance to the irrelevant content in the inputs. Moreover, the predictions from the HP model are used to enrich the inputs to the AR model during both training and testing. Experimental results show that our proposed model is superior to existing models for action recognition on the HMDB-51 dataset and the Penn Action dataset.

*Index Terms*—Action Recognition, Dense Correspondence Labelling, Domain Discrepancy, Task-irrelevant Content.

## I. INTRODUCTION

Action recognition has achieved an increasing popularity in the field of computer vision due to its potential applications in person identification, behaviour analysis, surveillance and recommendation systems. Hand-designed features such as improved Dense Trajectory (iDT) [1], Dense Trajectory (DT) [2], Trajectory-Pooled Deep-Convolutional Descriptors (TDD) [3] and deep learning-based models [4] [5] [6] [7] [8] have contributed to the considerable progress in video action recognition.

The first type of models considers action recognition as a classification task and conduct frame-wise predictions before aggregating the results of different frames. [9] samples one image from each video. Temporal Segment Networks (TSN) proposed in [8] samples 3 to 7 images per video before averaging the corresponding predictions. Long Short-Term Memory (LSTM) has also been widely applied to action recognition [10] [11] [12] [13]. For example, [11] proposed to sum up frame-level predictions by connecting LSTM cells to the output of underlying CNNs which conduct frame-level action recognition. However, a partial observation of videos may easily result in a loss of discriminative features.

The second type of existing work explores a distinguishing ingredient for action recognition, i.e. the temporal features. Comprehensive types of temporal information have been explored in [14] and [15] which obtain long-range temporal information with 3D CNNs. However, the performance of 3D CNNs and 3DHOG [16] are inferior to that of optical flows. DTPP proposed in [17] utilized video-level representations in multiple temporal scales to fully represent motion dynamics and generate predictions based on both RGB images and optical flows. Additionally, [18] introduced a new type of feature representation based on the spatial and temporal gradients of feature maps. The representation is orthogonal to optical flow and is computationally more efficient.

However, none of the above-mentioned methods have paid attention to domain discrepancy. The first type of domain discrepancy exists between the training set and the test set while the second gap is among different datasets. The former hinders the performance of generalization and the latter requires one model to be trained for each dataset. For instance, the most widely used datasets UCF-101 [19] and HMDB-51 [20] only have about 130 videos per class which are taken under constrained environments. The durations of videos as well as the variations covered by the videos in each class are much less than those in Youtube-8M [21] which is less constrained and better preserves complex real-world conditions. Therefore, the learned representations are difficult to generalize well. However, datasets such as Youtube-8M [21] require a huge amount of computational resources. As a result, the research towards narrowing the domain discrepancy with limited data and computational resources is desired to cope with realistic challenges.

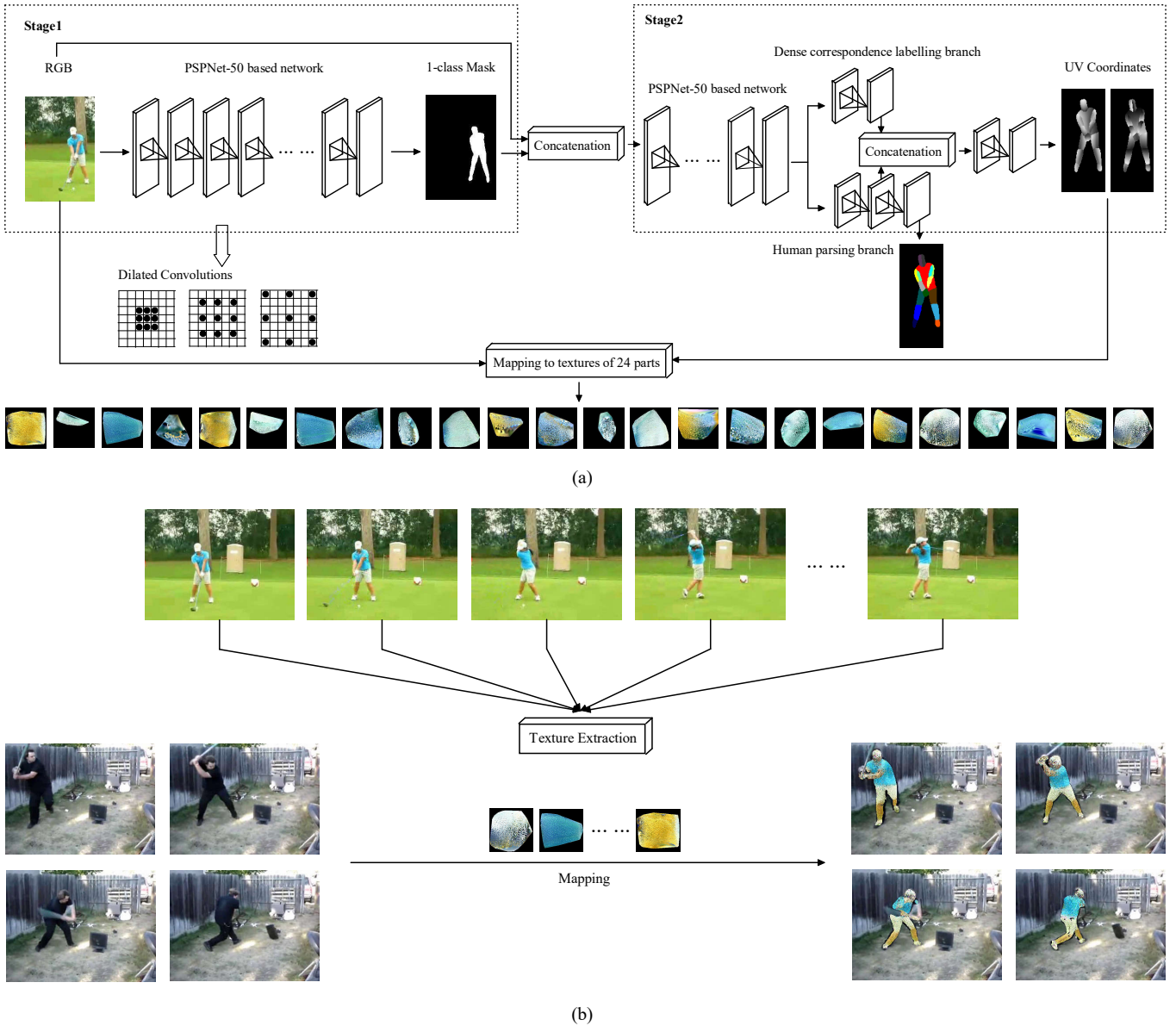

Fig. 1. Our proposed way of data augmentation based on texture transferring. (a) The model for mapping the pixels on a human image to the textures describing 24 semantic person parts. The UV coordinates introduced in [22] are used for dense correspondence labelling. The model is composed of two stages. The first stage performs saliency prediction while the second stage jointly predicts part segmentation masks and UV coordinates. (b) One single frame can only provide a partial view about the 24 parts. Therefore, we fill in the texture details using different views which describe the same set of parts.

To address the above-mentioned challenges and to facilitate a model towards reducing domain discrepancies, we propose a new way of determining irrelevant contents in input images. The contents have little mutual information with the labels and do not contribute to predictions, as will be discussed in Section III-B. To reduce the risk of overfitting caused by the irrelevant contents, an HP model for human parsing and densely localizing human textures is developed and is shown in Fig. 1 (a). The HP model performs almost as well as the state-of-the-art model with less parameters. It maps the pixels in an RGB image to a surface-based representation of human bodies and serves the purpose of re-rendering each video in the HMDB-51 dataset [20] using the textures collected from another randomly sampled video, as is illustrated in Fig. 1 (b). The collection of textures is based on the side views from different frames which offer complementary descriptions. By pairing the original videos with re-rendered ones for training, the AR model achieves invariance to human appearances and reduces the domain discrepancy.

Moreover, a procedure is proposed to concatenate the predictions from the HP model with RGB frames to feed into the AR model during both training and testing. The AR model has the same number of parameters as the state-of-the-art model DTPP [17] during testing but performs better.

The contributions of the paper are in three aspects: (1) A human passing model (HP model) is developed to conduct part segmentation and dense correspondence labelling. It achieves the similar performance as the current state-of-the-art model [23] and is used to collect textures from videos and enrich the HMDB-51 dataset [20]. (2) The dense correspondences predicted by the HP model serve the purpose of determining the task-irrelevant contents in inputs. The benefits will be shown in Section IV-D. (3) A procedure is proposed to enrich the inputs to AR model by integrating the predictions from the HP model. The advantages of the AR model will be shown in experiments.

The rest of this paper is organized as follows. Section 2 surveys the methods that are related to our work. Section 3 discusses our proposed HP model as well as the strategies for training and testing AR model. Section 4 explains our implementation details and reports and discusses experimental results. Concluding remarks are drawn in Section 5.

## II. RELATED WORK

**Conventional methods.** The most widely used hand-crafted features are DT-based and iDT-based features proposed in [1] and [2]. DT builds representations along the trajectories provided by optical flow estimations. iDT compensates the motion of cameras to remove the influences brought by shaking hands. iDT is also helpful in improving the performance of deep learning models, as was discussed in [17]. However, deep features perform much better than hand-crafted ones.

**Frame-level feature learning.** Early deep learning models for action recognition fused the predictions on single frames to obtain video-level predictions. For instance, [24] divided videos into fragments with different numbers of frames. The prediction in each fragment is based on a single-frame, and early fusion and late fusion were also proposed. [11] introduced different ways of temporal pooling for feature fusion and could process longer videos than those in [24]. [25] evaluated the importance of different frames using rank-pooling and proposed an efficient video-level descriptor. Similarly, [26] sampled 100 distinctive frames from every 450 ones and conducted average pooling. ECO was introduced in [27] which combined 2D convolutions with 3D convolutions for representing temporal and spatial relationships. However, frame-level feature representations are easier to suffer from over-fitting and cannot generalize as well as video-level ones without appropriate techniques for feature fusion. As a result, the long-range temporal features are required to characterize videos.

**ConvLSTM.** For the purpose of capturing the patterns which describe long-range temporal relationships in videos, Long Short-Term Memory (LSTM) module is integrated into an action recognition model to encode the relationships between different frames [28] [29] [11]. [11] and [30] proposed to use a 5-layer LSTM with 512 hidden nodes in each layer to aggregate the features from the RGB stream with those from the optical flow stream. [31] introduced the end-to-end Recurrent Pose-Attention Network (RPAN) where a spatial attention mechanism based on poses was used to integrate pose-relevant features in all frames into the feature representation of action recognition. This method facilitates an LSTM to learn the structure of movements along time. [32] applied temporal dilated convolutions which operated on multiple time steps to capture long-range temporal information. [33] proposed to sample task-relevant frames and aggregated local features before classification. This mechanism is similar to temporal attention. However, LSTM or RNN-based architectures might bring additional parameters which increases the computational burden and the risk of over-fitting. Our proposed scheme increases the robustness of a model without increasing the number of learnable parameters.

**3D Convolutions.** To make use of long-range temporal information without increasing complexity, 3D CNNs are proposed. Different from LSTMs, each input is composed of multiple frames. As a result, the temporal information can be captured without segmenting frames into snippets [4] [34] [15] [35] [14]. Early research on 3D CNNs include [36] which combined the features from an RGB stream, an optical flow stream and a stream of gradients. [15] introduced the C3D framework which improved the efficiency of 3D CNNs. [37] decomposed each 3D convolution into a spatial 2D operation and a temporal 1D operation. I3D proposed in [4] integrated the two convolutional operations in a two-stream architecture into a 3D one. Moreover, the spatial and temporal streams in [4] were trained independently. [38] and [39] proposed "P3D ResNet" to efficiently decompose 3x3x3 convolutions into 1x3x3 ones for spatial feature extraction and 3x1x1 temporal ones for temporal feature extraction. [40] proposed a SMART block to learn spatial and temporal features independently. [41] demonstrated that pre-training on the Kinetics dataset improves the performance of generalization. [42] improved conventional 3D CNNs by increasing the sizes of receptive fields, and the activations from spatial and temporal streams were fused to produce responses.

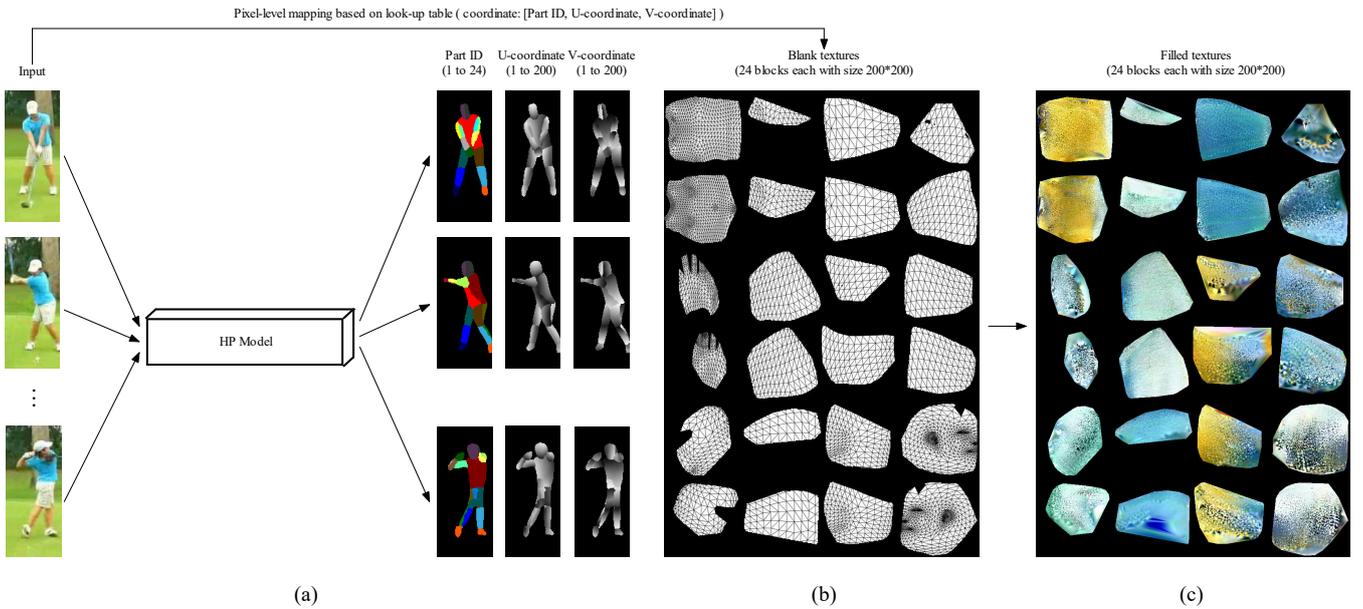

Fig. 2. The procedure for sampling textures from continuous frames. Left column: input frames illustrating a person from multiple point-of-views. [PartID, U-coordinates, V-coordinates] – the predictions from the HP model shown in Fig. 1 (a), and the UV coordinates map the pixels on images to the 24 look-up tables of textures, each one of the 24 sets of textures including a 200*200 block. Right column: the look-up table filled by the textures collected from input frames. Each frame provides one specific point-of-view of the same person.

However, the performance of 3D CNNs is hindered by two disadvantages: (1) The motion representations developed are not as good as those learned by a two-stream method based on optical flows. (2) A 3D CNN is quite memory-consuming because it takes all frames as inputs, as was reported in [17]. Our proposed scheme is based on a two-stream structure.

**Two-stream methods.** A framework with a two-stream architecture can be trained with offline-computed data such as optical flow estimations or iDT information. [6] demonstrated that concatenating the convolutional features from two branches using 1x1 convolutions contributes to better performance. To reduce the risk of over-fitting in the spatial stream, [43] proposed to multiply the feature representations from two branches. [8] introduced TSN which removed similar frames through dividing videos into snippets and uniformly sampling from snippets. The integration of RGB images, optical flows as well as warped optical flows was also implemented. [44] proposed ActionVLAD which jointly encoded both spatial and temporal features after aggregating them. However, the dimension of ActionVLAD representation is with size 32,768. It is of low efficiency in implementation. Moreover,

[45] built a temporal conditional random fields (CRFs) for making predictions based on multiple aspects of actions. [46] implemented temporal attention mechanism by detecting critical events from videos. End-to-end spatiotemporal networks were proposed to fuse both types of features using a pyramid architecture [47] [5] [48] [49]. Temporal attention based on a compact bilinear layer and spatial attention based on a spatiotemporal attention module are combined in the architecture. [50] reduced the redundancy from both data and CNNs' structures. However, the two-stream structure suffers from low efficiency. As a result, we propose to enhance the performance of a CNN model by upgrading the strategy of training. This method does not add to the number of learnable parameters.

III. METHODOLOGY

The work is divided into two parts: The first part involves the development of an HP model for human parsing and dense correspondence labelling. The predicted correspondences map colored pixels in images to the locations of textures in different person parts. The collections of textures from videos are based on the correspondences. Different from conventional models for pose estimation, the HP model also provides the point-of-view of each semantic part. For instance, even if the same semantic part appears in two frames, the first frame may show its frontal view while the second frame shows its side view.

The second part includes the analysis about the task-irrelevant contents in input images. Moreover, the re-rendered videos are generated for improving the robustness of the AR model during training. Finally, the method for enriching the inputs to the AR model is proposed.

*A. Nested CNNs for human parsing and texture extraction*

We name this model as HP model. The predicted correspondences map colored pixels on human images to the textures describing the 24 semantic person parts. Different from two-stage methods such as Dense-Pose [23], Faster-RCNN [51] and Mask-RCNN [52] which involve over 100 layers in feature extractors, our proposed HP model has much less parameters. As a result, the HP model spends less time during both training and testing.

The first sub-net conducts two-class semantic segmentation (humans and backgrounds) and it serves as a prior to the second stage, as is illustrated in Stage 1 of Fig. 1 (a). We leverage the PSPNet-50 [53] as the backbone. The prediction is formulated as $\mathbb{R}^{H \times W \times C} \mapsto \mathbb{R}^{H \times W \times C'}$ where $H$ and $W$ denote, respectively, the height and width of input and output images with $C=3$ and $C'=1$. The pixel-wise cross-entropy loss function is computed based on the binary pixel-wise annotations for training stage 1.

The second sub-net concatenates the original RGB images with the predictions from the first sub-net as the inputs. It estimates a fine-grained semantic parsing map with 24 labels and UV coordinates for dense correspondences. The predictions are formulated as $\mathbb{R}^{H \times W \times (C+C')} \mapsto \mathbb{R}^{H \times W \times C''}$ where $C''=3$. As is shown in Fig. 1 (a) and Fig. 2, each output has 3 channels: PartID, U-coordinates and V-coordinates. PartID maps each pixel to its corresponding part and (U-coordinates, V-coordinates) is the exact location of the pixel within the part using the look-up table corresponding to that part. Each part is parameterized by a unique look-up table with size $200 \times 200$, as is illustrated in Fig. 2 (b) (c). Stage 2 has the same backbone as Stage 1. Two heads are built on top of the backbone, one for part segmentation and the other for coordinate regression.

Different from [23] which utilized the available textures from the SURREAL dataset [54], we generate textures using color information from video clips in the HMDB-51 dataset [20]. The procedure is demonstrated in Fig. 2 and is based on the SMPL format of dense correspondences [22]. 24 200x200 planes compose one set of textures describing one person. Pixels in RGB frames are mapped to the planes with relative locations kept. As there are only limited pixels available, interpolation is conducted to in-paint the texture details that cannot be collected from video frames. The filled textures are shown in Fig. 2 (c). As is shown in Fig. 2 (a), people with different poses can provide the details of different parts or different components of the same part. For instance, the frontal pose shows the textures of chest while the view of the person's back shows the details of back. Only by collecting the textural details from different views can the dense correspondences of all 24 parts be obtained.

*B. Input content selection based on dense correspondences*

In this section, we evaluate the correlation between human appearances and the task of action recognition. Different from the procedures for feature selection in machine learning [55], we directly evaluate the redundancy in inputs and remove the contents in the input data which are irrelevant to the target task.

[55] divided the selection of input features into two steps: subset search and subset evaluation based on Eq. (1).

$$Gain(A) = Entropy(D) - \sum_{v=1}^{V} \frac{|D^v|}{|D|} Entropy(D^v) \quad (1)$$

where $D$ denotes the dataset and $A$ indicates the subset of features under evaluation. $D$ is divided by $A$ into $V$ subsets each of which share the same value on $A$. $Gain(A)$ indicates the contribution made by $A$ to correctly classifying $D$. Suppose that $D$ has $N$ class labels and the proportion of input samples belonging to the $i$-$th$, $1 \leq i \leq N$ class is $p_i$.

$$Entropy(D) = \sum_{i=1}^{N} p_i \log \frac{1}{p_i} \quad (2)$$

Similarly, we firstly divide input images into contents and then evaluate the relevance of human appearance contents to the action recognition task. The dense correspondences between RGB images and the surface-based representations of 24 human parts provided by the HP model in Fig. 1 (a) enables us to evaluate the contribution from human appearances to classification. Firstly, Eq. (1) is reformulated as

$$\begin{aligned} Gain(X) &= \sum_y p(Y=y) \log \frac{1}{p(Y=y)} - \\ &\quad \sum_x p(X=x) \sum_y p(y|X=x) \log \frac{1}{p(y|X=x)} \quad (3) \\ &= H(Y) - \sum_x p(X=x) H(Y|X=x) \\ &= H(Y) - H(Y|X) \end{aligned}$$

where $X$ represents an input sample containing only human appearances and $Y$ indicates a ground truth label in action recognition. The first term in Eq. (3) denotes the entropy of the dataset. The second term measures the remaining uncertainty in the dataset if provided $X$. $Gain(X)$ measures the contribution made by $X$ to the accurate prediction of $Y$. In another way, the mutual information between random variables $X$ and $Y$ with joint distribution $p(x,y)$ can be expressed as

$$\begin{aligned}I(X;Y) &= D_{KL}\big[p(x,y)\|p(x)p(y)\big]\\&=\sum_x\sum_y p(x,y)\log\left(\frac{p(x,y)}{p(x)p(y)}\right)\\&=\sum_x\sum_y p(x,y)\log\left(\frac{p(y|x)}{p(y)}\right)\\&=H(Y)-H(Y|X)\end{aligned} \quad (4)$$

As a result, the relevance of $X$ to the target task with the label $Y$ is equivalent to the mutual information between $X$ and $Y$. Under the assumption that a deep learning model is best at extracting the features in inputs that are informative about labels, we evaluate $I(X;Y)$ by training a CNN model for predicting $Y$ given $X$ and compare the performance of predicting $Y$ using $X$ and randomly guessing $Y$.

To encode only the information of human appearances, our AR model is trained on the inputs which only contain the textures of 24 human parts. The images are with size 1200-by-800 and an example is illustrated in Fig. 2 (c). The setting up is shown in Fig. 3.

The accuracy over the training process will be shown in Section IV-D.

### C. The method for training AR model

The AR model has the RGB stream and flow stream discussed in [17]. The backbone is based on the BN-Inception Network [56]. The difference between the RGB stream in our AR model and that in DTPP Model [17] lies in the procedure of training, as is illustrated in Fig. 4.

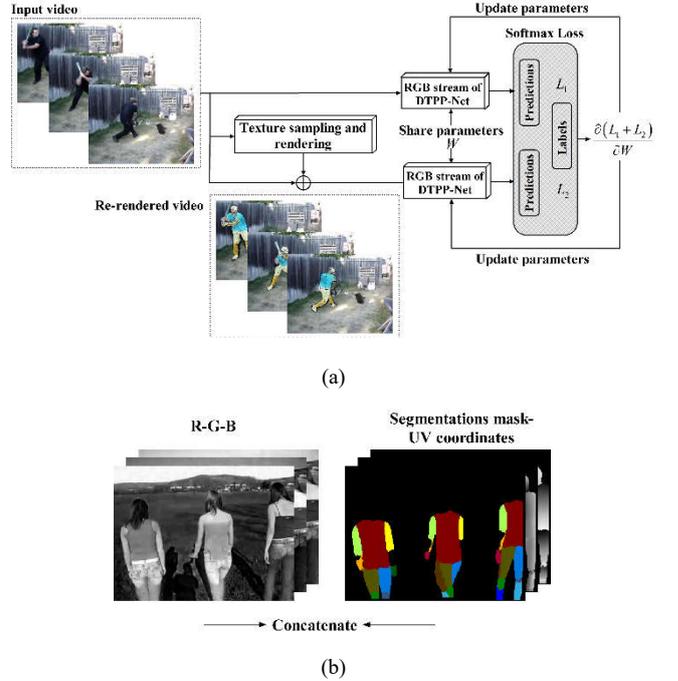

(a)

(b)

Fig. 4. The procedure for training our AR model. (a) Each input video is re-rendered using a set of textures which are collected from another randomly sampled video from the HMDB-51 dataset. The procedures for texture rendering have been illustrated in Fig. 1 (b) and Fig. 2. The network inherits its structure from the DTPP-Net proposed in [17]. Paired images are located in the same batch to facilitate the averaging of gradients. (b) After being pre-trained according to (a), the AR model is trained for a second time. Each input frame to our AR model is composed of 6 channels which is the concatenation of the original RGB and the predictions from the proposed HP model. Each input frame is composed of 6 channels during both training and testing.

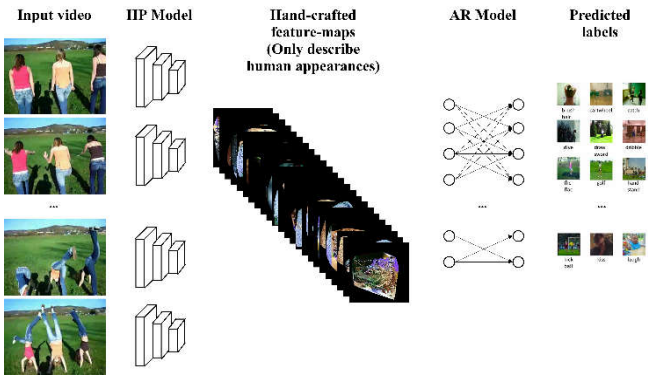

Fig. 3. The AR model is re-purposed to conduct single-frame classification. The feature-maps with 24 channels are predicted by the HP model, describing the appearances of human parts only. Each channel has a size of $200 \times 200$. The AR model learns to predict action labels based on the feature-maps.

Firstly, the humans in all input videos are re-rendered using the method shown in Fig. 1 (b). The source videos for extracting textures and re-rendered target videos are randomly paired. The first branch in Fig. 4 (a) corresponds to an original video from HMDB-51 dataset while the second one shows the re-rendered version. Both branches share everything but the human appearances in inputs. The second branch only exists during training. The batch size for training is 4 and two pairs of videos are involved in each batch. The gradients from both the original branch and the re-rendered branch are averaged to update the parameters. Although the inputs are different, the two branches are forced to learn the same feature representations which is invariant to the changes of appearances.

The procedure for training is divided into two steps, as is described in Algorithm 1. In the first step, the AR model is trained with 3 input channels, as illustrated in Fig. 4 (a). Then each input frame is expanded to be with 6 channels and The AR model is fine-tuned on the expanded frames.

| Algorithm 1. The procedure for training | |
| --- | --- |
| Steps | Operations |
| 1 | Pair each video clip from the HMDB-51 dataset with its 9 re-rendered versions. Train the spatial stream of the AR model with pairs of videos according to Fig. 4 (a). Each batch is composed of two pairs of videos. |
| 2 | Collect the prediction of the HP model on each frame and concatenate the prediction with its corresponding RGB frame. Modify the first convolutional layer of the spatial stream and train it on new inputs with 6 channels. |

This training procedure has two merits. Firstly, the AR model develops a robust feature representation and the training instances in the learned manifold have lower intra-class distances. The model achieves invariances to different human appearances which are irrelevant to action recognition. As a result, the learned feature extractors tend to concentrate on more discriminative features, especially when faced with confusing data. The domain discrepancy between training data and test data is reduced and performance is improved. Secondly, involving segmentation masks as well as coordinates of textures as the inputs provides the descriptions of both poses and the views of semantic parts. For instance, the frontal view and back view of the same semantic part are distinguished by different textural coordinates within the same part. A comparison between the proposed AR model and existing benchmarks will be presented in Section IV-E and Section IV-F.

## IV. EVALUATION

### A. Introduction to Datasets

The HP model and the AR model were trained on different datasets. At Stage 1, the HP model was trained on the union of the MHP v2.0 dataset [57] and the LIP dataset [58]. At Stage 2, the HP model was trained on the Dense-pose COCO dataset [23]. The AR model was trained and tested on the HMDB-51 dataset [20].

**MHP v2.0 Dataset.** The dataset contains 25,403 images with 58 fine-grained semantic category labels. It is divided into 3 splits: 15,403 for training, 5,000 for validation, and 5,000 for testing. In our experiments, the labels on all semantic person parts are assigned the same class label while the backgrounds are assigned the other label. The evaluation metric for this dataset includes Average Precision based on parts ($AP^p$), Average Precision based on regions ($AP^r$) as well as Percentage of Correctly parsed Parts ($PCP$). $AP^p$ computes the average of Intersection over Union ($IoU$) of all part categories. $AP^r$ evaluates the average of $IoU$ of all instances. $PCP$ evaluates the proportion of parts that are parsed with an $IoU$ over a certain threshold. In our implementation, $AP^p$ is the same as $AP^r$ because there is only one foreground category. The loss function of Stage 1 in HP model is based on $AP^r$.

**LIP Dataset.** The dataset contains 50,462 images with pixel-wise annotations covering 19 semantic person part labels. We have converted the 19 labels into 1 uniform foreground label. The images are with challenging poses and views, heavily occlusions, various appearances, and low resolutions. The metric for evaluation is the same as that in the MHP dataset.

**Dense-pose COCO Dataset.** The HP model was trained on the Dense-pose COCO dataset proposed in [23]. The dataset involves the annotations for 50k humans with over 5 million manually annotated correspondences in the form of UV coordinates. The training set contains 48k humans and the test set has 1.5k images with 2.3k humans. The metric for evaluating performance on this dataset includes Area Under the Curve (AUC) and $IoU$.

**HMDB-51 Dataset.** The dataset consists of realistic videos from multiple sources, including movies and web videos. It is composed of 6,766 video clips which cover 51 action categories. Standard 3 training/testing splits are used for evaluation. The evaluation metric adopted for this dataset is classification accuracy (%). We have re-rendered each video with 9 sets of textures which are randomly sampled from 9 other videos. The augmented dataset contains 67,660 video clips.

**Penn Action Dataset.** The dataset was proposed in [59] and includes 2,326 videos from 15 categories of actions. For instance, "baseball_swing", "pull_ups" and "strumming_guitar" are three of these classes. The videos were obtained from multiple online sources. The number of frames in each video ranges from 18 to 663. The frames in this dataset suffer from occlusions as well as significant variances in scales. Experimental results on this dataset are presented in Part E.

### B. Implementation Details

**HP Model.** Different from both the end-to-end Fully Convolutional Network (FCN) structure and the two-stage RCNN structure [23], the two stages of the HP model can be trained together and the second stage takes both the heat-maps and the original RGB images as the inputs. As is shown in Fig. 1 (a), Stage 2 is built on Stage 1 and conducts refined regression based on the predictions of Stage 1. The training of HP model is divided into two steps. Firstly, at Stage 1, it is trained on itself with the initial learning rate set to 2.5e-4 and the input images cropped to 473-by-473. After the validation accuracy saturates, we keep the learnt parameters from Stage 1 and concatenate the output heat-maps with the original images as the inputs of Stage 2. At Stage 2, the model is trained on itself to map the inputs to semantic part labels and UV coordinates. The part labels and UV coordinates are utilized as discussed in Fig. 1 (b) to collect textures and re-render videos. Note that the videos from the classes such as "brush_hair", "chew", "eat", "drink", "dribble", "kiss", "laugh", "smile", "smoke", and "talk" part are not re-rendered because the details of faces cannot be perfectly described by textures. The batch size was set to 12 for training. The implementation was based on the Caffe platform [60] with two NVIDIA GEFORCE GTX 1080 Ti GPUs.

**AR Model.** For the HMDB-51 dataset, each video was split into 25 segments during training and testing, i.e. 25 frames were sampled from each video. For the Penn Action dataset, 16 frames were sampled from each video clip. Both the spatial stream and the temporal stream were trained with the categorical soft-max loss. The batch size was set to 4. The RGB stream and optical flow stream were trained on two GPUs independently and tested together. The learning rate was initialized to 0.01 and was divided by 10 each time when validation loss saturates until 0.00001. The momentum was 0.9 and gradients were clipped at a L2-norm of 40. The dropout ratio was set to 0.8. As is illustrated in Fig. 4, we have paired each training video with its 9 re-rendered versions and each batch contains 2 video pairs. The implementation was also based on the Caffe platform [60] and two NVIDIA GEFORCE GTX 1080 Ti GPUs were used.

### C. Evaluation of HP model

The metric for evaluating the accuracy of predictions on UV-coordinates is AUC which is computed by

$$AUC_a = \frac{1}{a}\int_0^a f(t)dt \quad (5)$$

Two different values of $a$ ($10cm$ or $30cm$) are adopted to measure how many points are estimated with an error of less than 10 cm or 30 cm. The AUC for the $j-th$ estimated point is computed by

$$Distance(j) = \exp\left(\frac{-g(i_p,\hat{i}_p)^2}{2\kappa^2}\right) \quad (6)$$

The details in human images are described by vertices using the SMPL model introduced in [22]. $i_p$ denotes the location of the vertex that is closest to the ground truth location of the $j$-th point and $\hat{i}_p$ locates the vertex that is closest to the $j$-th estimated point. $g(\cdot,\cdot)$ measures the geodesic distance between two points on a surface. $\kappa$ is set to 0.255 when $Distance(j) = 0.5$ corresponds to $AUC(j) = 30cm$. $\kappa$ is set to 1 when $Distance(j) = 0.96$ corresponds to $AUC(j) = 30cm$. $\kappa$ is set to 1.45 when $Distance(j) = 0.5$ corresponds to $AUC(j) = 170cm$. In our experiments $\kappa$ was set to 0.255. Table 1 shows a comparison on performance. Intersection over Union (IoU) proposed in [23] is used for evaluating the performance of human parsing.

Table 1. Performance comparison on dense correspondence labelling. The performance of our HP model is compared with that of current state-of-the-art DensePose RCNN and DensePose FCN [23] on the COCO Densepose dataset [23]. The images in the dataset contain multiple people with variances in poses, backgrounds and scales.

| Methods | $AUC_{10}$ | $AUC_{30}$ | IoU |
| --- | --- | --- | --- |
| DensePose-FCN | 0.253 | 0.418 | 0.66 |
| DensePose-RCNN (only points) | 0.315 | 0.567 | 0.75 |
| DensePose-RCNN (distillation) | 0.381 | 0.645 | 0.79 |
| DensePose-RCNN (cascade) | 0.390 | 0.664 | 0.81 |
| HP model | 0.383 | 0.651 | 0.80 |
| Human Performance | 0.563 | 0.835 | |

From Table 1, we can see that our proposed HP model performs almost as well as the current state-of-the-art model (DensePose-FCN and DensePose-RCNN [23]) with only 52.7% as many parameters.

*D. Determining the task-irrelevant content using the predictions from the HP model*

The accuracy over the training process is shown in Fig. 5. It can be observed that the probability of labels provided by appearances $p(Y=y|X=x)$ is almost equal to $p(Y=y)$ because the number of classes is 51. A probability of $p(Y=y)=0.0196$ can be achieved by random sampling without any information from $X$. According to Eq. (4), $I(X,Y)$ approximates zero.

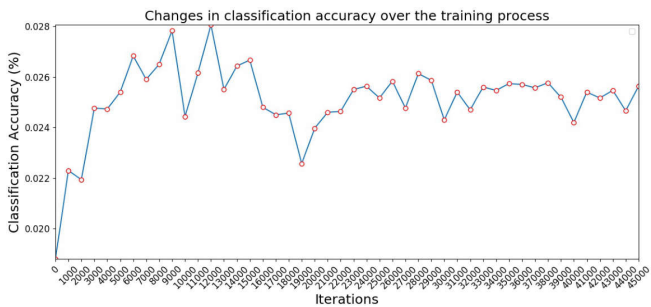

Fig. 5. Training classification accuracy of the AR model shown in Fig. 3. The training was conducted for 45,000 iterations without obvious increase in accuracy.

To further demonstrate that human appearance is irrelevant to action recognition, we replace the human regions in images with the predictions from the HP model. In that way, humans in the videos of all classes are with the same appearance. The revised images are used during training and testing. Fig. 6 shows an example.

The AR model is trained and tested on the revised images shown in the bottom rows in Fig. 6 (a) and Fig. 6 (b). The results in Table 2 demonstrate the effectiveness of removing task-irrelevant contents.

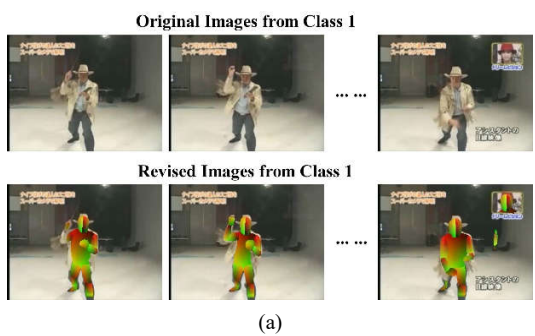

(a)

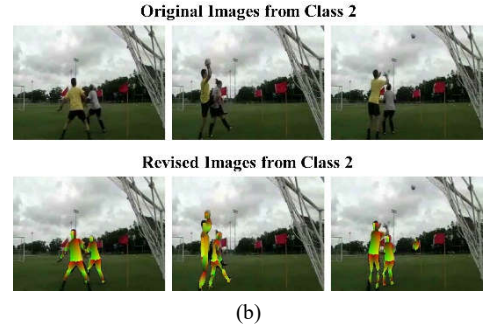

(b)

Fig. 6. The human regions are replaced with the outputs from the HP model. (a) and (b) show the frames from two videos which have different class labels in the HMDB-51 dataset. Upon the revision, humans in videos of different classes are with the same appearance.

Table 2. The effect of revising training and testing images on the performance of the AR model. The experiment is conducted on the three testing splits of HMDB-51. Performance is measured in classification accuracy (%).

| Methods | Split 1 | Split 2 | Split 3 |
|---|---|---|---|
| Spatial Stream (AR model) trained and tested on the original images in HMDB-51. | 61.5 | 61.2 | 60.5 |
| Spatial Stream (AR model) trained and tested on the revised images. | 62.8 | 62.7 | 61.6 |

From Table 2 it can be observed that revising the humans in all videos to be with the same appearance does not lead to any decrease in accuracy. The method of enforcing the invariance to human appearances will be conducted in Part E.

*E. Evaluation of AR model*

The performance of the AR model is evaluated on the HMDB-51 dataset. During both training and testing, 25 frames were sampled from each video, and the scores from the temporal stream and spatial stream were fused. A temporal pooling layer of 3 levels proposed in [17] was applied in both streams. The procedure for learning appearance-invariant representations and enriching inputs was applied. As the AR model is a modified version of the temporal stream in DTPP which is proposed in [17], we compare the AR model with DTPP in Table 3.

The 6,766 videos are divided into three splits. The first split has 3,570 videos for training, 1,531 videos for validation and 1,665 videos for testing. The second split has 3,570 videos for training, 1,532 videos for validation and 1,664 for testing. The third split has 3,570 videos for training, 1,532 videos for validation and 1,664 for testing. According to the strategy of data augmentation and pairing, we expanded the three splits to be with 357,00 training videos and 15,310 validation videos, 357,00 training videos and 15,320 validation videos, 357,00 training videos and 15,320 validation videos, respectively. The results are shown in Table 3. Ablation study was conducted to evaluate the improvements brought by each of the two steps in Algorithm 1 (Table 3).

Table 3. Performance comparison between the AR model and the DTPP Model [17] on the three testing splits of HMDB-51, measured in classification accuracy (%).

| Methods | Split 1 | Split 2 | Split 3 |
|---|---|---|---|
| Spatial Stream (DTPP) | 61.5 | 61.2 | 60.5 |
| **Spatial Stream (AR model) trained according to Fig. 4 (a)** | 61.9 | 61.7 | 60.9 |
| **Spatial Stream (AR model) with 6 input channels shown in Fig. 4 (b) and trained according to Fig. 4 (a)** | 63.4 | 63.2 | 62.4 |
| Temporal Stream (DTPP) | 66.3 | 69.2 | 68.8 |
| Temporal Stream (AR model) | 66.3 | 69.2 | 68.8 |
| Fusing two streams (0.4 for spatial, 0.6 for temporal) (DTPP) | 75.0 | 75.0 | 74.4 |
| **Fusing two streams (0.4 for spatial, 0.6 for temporal) (AR model trained using the techniques in Fig. 4 (a) and (b))** | 76.4 | 76.5 | 75.7 |
| Fusing two streams with MIFS (DTPP) | 76.9 | 76.3 | 75.9 |
| **Fusing two streams with MIFS (AR model trained using the techniques in Fig. 4 (a) and (b))** | 77.8 | 77.2 | 76.8 |
| Fusing two streams with iDT (DTPP) | 76.3 | 74.6 | 75.1 |
| **Fusing two streams with iDT (AR model trained using the techniques in Fig. 4 (a) and (b))** | 77.0 | 75.4 | 75.8 |

Two conclusions can be drawn from Table 3. Firstly, Step 1 of Algorithm 1 enables the AR model to develop the feature representations which are invariant to different human appearances. A slight improvement of about 0.5% demonstrates that the invariance to human appearances reduces the domain discrepancy. This conclusion is consistent with the discussion in Section III-B and Section IV-D. Besides developing the invariance to irrelevant contents, Step 2 in Algorithm 1 also enriches inputs with the predictions from the HP model. The performance is further improved. As a result, the AR model outperforms the DTPP Model [17] both in the single RGB stream and in the fusion of both streams. Note that the MIFS model [61] stacks the features which are extracted using a family of differential filters parameterized with multiple time skips, while the iDT model [1] cancels out camera motions from both streams. The predicted scores from MIFS and iDT are available online [17]. Besides the first convolutional layer which involves 6 input channels, the AR model has the same architecture as the DTPP Model during testing.

The AR model is also compared with several other benchmark models. The average accuracy of different models on the three splits of the dataset is shown in Table 4.

Table 4. Performance comparison of AR model with existing benchmarks.

| Models | Classification Accuracy on HMDB-51 (%) |
|---|---|
| iDT [1] | 57.2 |
| MoFAP [62] | 61.7 |
| MIFS [61] | 65.1 |
| Two-stream [9] | 59.4 |
| TDD [3] | 63.2 |
| FstCN [37] | 59.1 |
| LTC [14] | 64.8 |
| TSN (3 segments) [8] | 70.7 |
| TSN (7 segments) [8] | 71.0 |
| DOVF [33] | 71.7 |
| ActionVLAD [44] | 66.9 |
| ST-ResNet [6] | 66.4 |
| ST-Multiplier [43] | 68.9 |
| ST-Pyramid Network [47] | 68.9 |
| TLE [5] | 71.1 |
| Four-Stream [63] | 72.5 |
| **DTPP** [17] | 74.8 |
| ST-ResNet + iDT [6] | 70.3 |
| ST-Multiploer + iDT [43] | 72.2 |
| DOVF + MIFS [33] | 75.0 |
| ActionVLAD + iDT [44] | 69.8 |
| Four-Stream + iDT [63] | 74.9 |
| **Our proposed AR model** | 76.2 |

From Table 4, we can see that our AR model outperforms existing benchmark methods even without the augmentation of MIFS or iDT. To have a better comparison, Fig. 7 shows that our AR model outperforms the DTPP model on most of all 51 classes in HMDB-51 dataset [20].

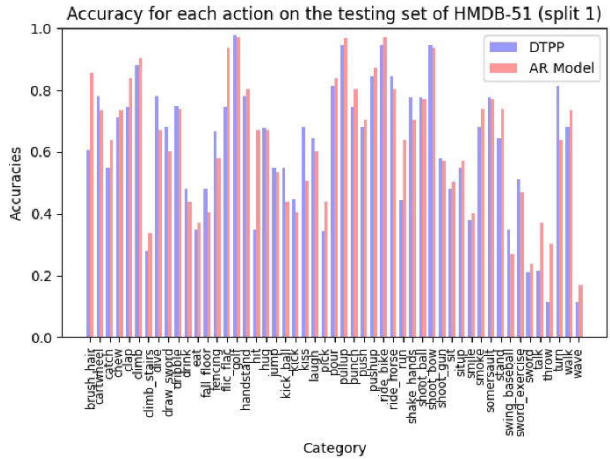

Fig. 7. Classification accuracy for each action class on the test set of HMDB-51 dataset [20].

*F. Evaluation on Additional Data for Action Recognition*

It has been shown in Fig. 4 (a), (b) and Table 3 that the integration of predictions from HP model contributes to improvements in action recognition. However, the dataset is not large enough to cover all types of cases. Therefore, the proposed model is also evaluated on the Penn Action dataset [59] to show its benefits.

Similar to the scenario in HMDB-51 dataset, the HP model was firstly applied to build the dense correspondences between the pixels in RGB frames from Penn Action dataset [59] and the surface-based representation of human bodies. Then optical flow estimation was conducted on the frames provided in the Penn Action dataset. The spatial and temporal streams of the AR model were trained on RGB frames and optical flow frames, respectively.

The frames in each video are divided using a 50/50 training/testing split. Performance is measured in classification accuracy and is shown in Table 5.

Table 5. Classification accuracy on the Penn Action dataset.

| Models | Accuracy (%) |
|---|---|
| Joint action recognition and pose estimation [64] | 85.5 |
| Pose for action [65] | 92.9 |
| Body joint guided method [66] | 95.3 |
| **Our model trained using the techniques in Fig. 4 (a) and (b)** | 96.8 |

From Table 5, it can be observed that our proposed method outperforms existing benchmark models for action recognition on the Penn Action dataset [59].

## V. CONCLUSION

In this paper, we propose the HP model for human parsing which conducts dense correspondence prediction and part segmentation. We also propose a novel strategy for training and testing the AR model. The HP model performs almost as well as the current state-of-the-art model with a much simpler training scheme and less parameters used. The proposed training strategy in Algorithm 1 reduces domain discrepancies by enforcing the invariance to irrelevant clues. Moreover, the predictions from the HP model enriches the input to the AR model by providing dense localizations of human parts and textures. The AR model benefits from the novel strategies and outperforms existing benchmark models with a higher robustness and more stable predictions.